\documentclass{article}
\usepackage{spconf,amsmath,graphicx}
\usepackage{comment}

\usepackage{multirow}
\usepackage{booktabs}
\usepackage[table,xcdraw]{xcolor}
\usepackage{cite}
\usepackage{url}
\usepackage{amsfonts}
\usepackage{bbm}
\usepackage{bm}

\title{Continual Learning with Embedding Layer Surgery and Task-wise Beam Search using Whisper}

\name{Chin Yuen Kwok$^1$, Jia Qi Yip$^{1,2}$ and Eng Siong Chng$^1$}
\address{$^1$ College of Computing and Data Science, Nanyang Technological University, Singapore\\$^2$ Alibaba Group, Singapore\\{\footnotesize\url{kwok0062@e.ntu.edu.sg, jiaqi006@e.ntu.edu.sg, aseschng@ntu.edu.sg}}}
\begin{document}
%\ninept
%
\maketitle
\begin{abstract}

Current Multilingual ASR models only support a fraction of the world’s languages. Continual Learning (CL) aims to tackle this problem by adding new languages to pre-trained models while avoiding the loss of performance on existing languages, also known as Catastrophic Forgetting (CF). However, existing CL methods overlook the adaptation of the token embedding lookup table at the decoder, despite its significant contribution to CF. We propose Embedding Layer Surgery where separate copies of the token embeddings are created for each new languages, and one of the copies is selected to replace the old languages embeddings when transcribing the corresponding new language. Unfortunately, this approach means LID errors also cause incorrect ASR embedding selection. Our Task-wise Beam Search allows self-correction for such mistakes. By adapting Whisper to 10 hours of data for each of 10 unseen languages from Common Voice, results show that our method reduces the Average WER (AWER) of pre-trained languages from 14.2\% to 11.9\% compared with Experience Replay, without compromising the AWER of the unseen languages.
\end{abstract}
\section{Introduction}
\label{sec:intro}

Recent advancements in training speech models involve utilizing millions of hours of multilingual ASR and translation labeled data, leading to the development of models supporting Massively Multilingual ASR (MMASR) which can transcribe over 50 languages \cite{pratap2020massively}. Additionally, newer models such as Whisper \cite{radford2023robust} and MMS \cite{pratap2023scaling} are language-agnostic \cite{datta2020language}, where a single model is able to perform both language identification (LID) and ASR seamlessly so that users do not need to manually specify the language they are transcribing. Nevertheless, even the most diverse MMASR models are not able to support the vast number of languages spoken globally \cite{katzner2002languages}, and the additional requirements for language-agnostic MMASR increases the difficulty of adding support for new languages after the initial pre-training.

%How additional laguages are supported?
To expand ASR support to additional languages, a naive solution is to adapt MMASR using only the new datasets containing the new languages. However, this causes catastrophic forgetting (CF) \cite{zhai2023investigating}, where a model has degraded ASR performance on previously learned languages after adapting to new ones. Therefore, this necessitates Continual Learning (CL), a class of adaptation methods that aims to mitigate CF. There are numerous CL methods for model adaptation which includes Prototype-based \cite{janson2022simple}, Regularization-based \cite{kirkpatrick2017overcoming,aljundi2018memory,li2017learning}, Replay-based \cite{brignac2023improving,buzzega2020dark}, Optimization-based \cite{chaudhry2018efficient} and Dynamic-architecture-based \cite{mallya2018piggyback} methods.

\begin{figure}[]
  \centering
  \includegraphics[width=\linewidth]{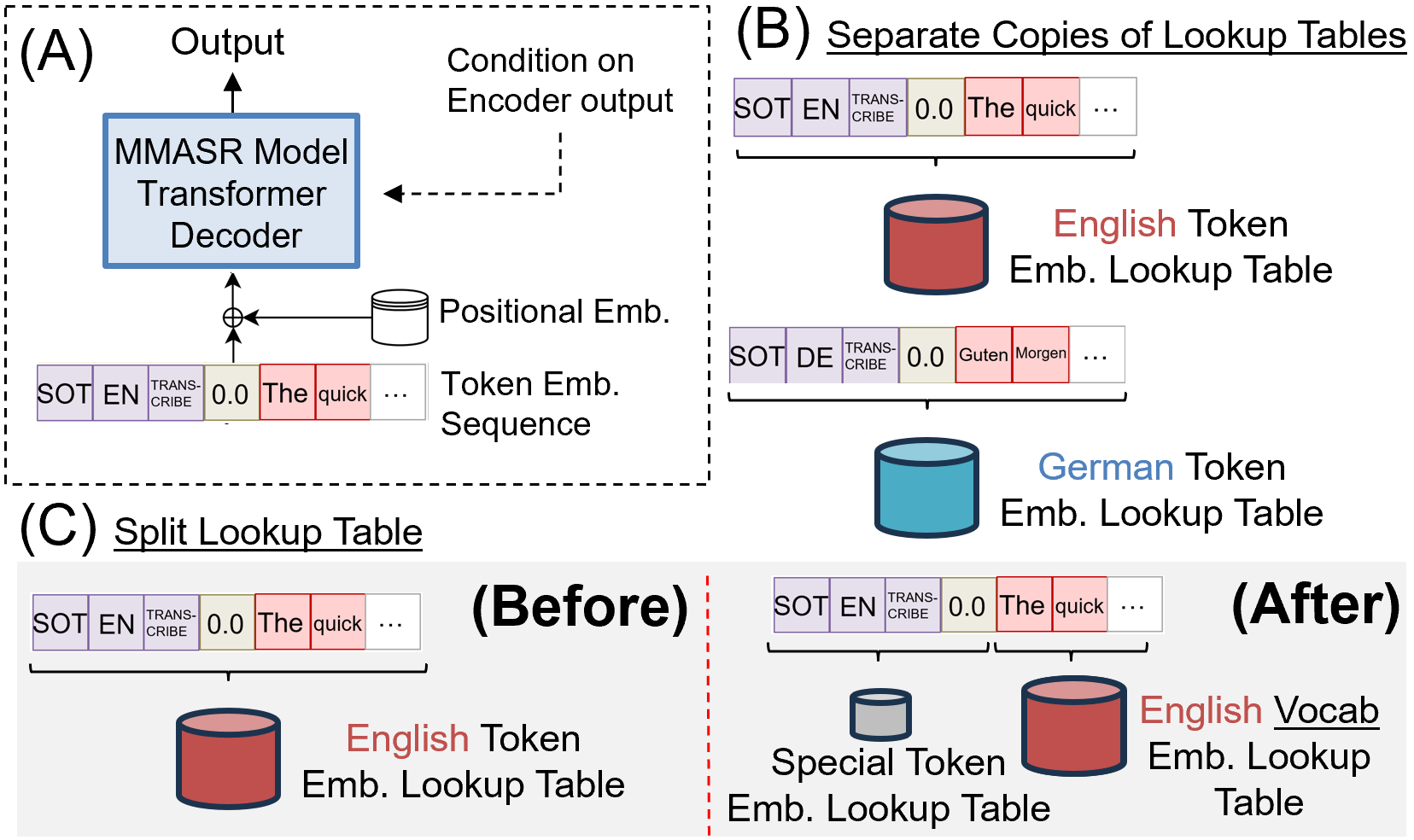}
  \caption{Overview of our Embedding Layer Surgery. (A) Token embeddings are fed to the MMASR model decoder. (B) A separate lookup table is used to generate token embeddings for each language. (C) The lookup table is further split into two parts. One for special tokens and one for vocabulary tokens.}
  \label{fig:split_emb}
\end{figure}

Specifically in the audio domain, prototype-based \cite{michieli2023online}, replay-based \cite{cappellazzo2023investigation}, and regularization-based \cite{vander2023rehearsal} methods are examined. In addition, model averaging \cite{vander2023rehearsal,yang2023dual} and layer-wise regularization \cite{wangclrl} approaches have also shown promise in CL for audio. However, limited focus has been given to CL for MMASR. While some works has shown success in mitigating CF by incremental fine-tuning \cite{li2022massively} , training a mapping matrix for task-specific weights \cite{yu2023master}, and performing weight averaging \cite{eeckt2022weight}, they are not language-agnostic. Another work \cite{della2023cl} has provided a comprehensive study of CL baseline methods for MMASR, indicating potential areas for improvement. 

Notably, existing MMASR CL approaches overlook adaptation of the token embedding lookup table \cite{huang2021lookup} at the decoder, because they mostly focus on non-language-agnostic MMASR models. However, the lookup table is important in the context of language-agnostic MMASR CL, as it contains the additional language ID token embeddings to perform LID. After adapting the model to new languages, these tokens are highly biased towards identifying any input audio as one of the new languages, causing CF in LID of the pre-trained languages. A second problem is that language-agnostic MMASR heavily relies on the accuracy of LID, as an LID error may cause the model to transcribe in the wrong language.

%Contribution
% What is our proposal and how does it link to the two problems with CL for language-agnostic MMASR systems?
% What exactly is the research gap?
To address CF in the token embedding lookup table during CL, we propose Embedding Layer Surgery where separate copies of the token embeddings are created for each new languages, and one of the copies is selected to replace the old languages embeddings when transcribing the corresponding new language. This allows the embeddings for the existing languages to be maintained, ensuring that they will not be overwritten after adapting to new languages. In addition, task-wise beam search is proposed to address the error propagation from language confusion. Our contributions are three-fold. We show that 1) using language-specific token embeddings can reduce forgetting, 2) a model can remain language-agnostic by using our Embedding Layer Surgery, and 3) LID can be enhanced by task-wise beam search, which in turn improves language-agnostic MMASR performance.

\section{Proposed Method}
\label{sec:method}

We use Whisper \cite{radford2023robust} as the pre-trained MMASR model and adapt it to the ASR and LID datasets of the new languages. LID adaptation is needed as it allows the model to transcribe in a language-agnostic setting \cite{datta2020language}, where instead of manually prompting Whisper to control the transcribed language, the model can use LID to automatically determine what language it should be transcribing.

To prevent catastrophic forgetting (CF) on previously learnt (old) languages while adapting to new languages, Experience Replay (ER) is used where a subset of the datasets containing the old languages are mixed with the new language adaptation dataset so the model can rehearse on old languages ASR tasks.

Additionally, we propose to further mitigate CF by using a separate text embedding lookup table \cite{huang2021lookup} for each new languages and splitting the lookup table into two parts as shown in the bottom-right part of Figure \ref{fig:split_emb}. 

Finally, we propose task-wise beam search as shown in Figure \ref{fig:task_beam_search} to retain a list of candidates of different languages during beam search to enhance the ASR and LID performance in the langauge-agnostic setting.

\subsection{Whisper Model for Language-agnostic MMASR}
Whisper \cite{radford2023robust} is used as the pre-trained MMASR model for adaptation. It supports 75 languages and also supports LID. It uses the Transformer attention-based encoder-decoder architecture and decodes in an auto-regressive manner. We select this model because while \cite{rouditchenko2023comparison} shows that Whisper performs similarly with XLS-R \cite{conneau2020unsupervised} in terms of averaging both the seen and unseen languages WER, \cite{della2023cl} shows that Whisper surpasses WavLM \cite{chen2022wavlm} overall in MMASR CL and \cite{praveen2023language} shows that Whisper performs better than wav2vec 2.0 models \cite{babu2021xls} and WavLM \cite{chen2022wavlm} for LID. The Whisper decoder is preferred over the Large Language Models (LLM) based decoder \cite{tang2023salmonn} for ASR as the latter does not support LID to the best of our knowledge.

\begin{figure}[t]
  \centering
  \includegraphics[width=\linewidth]{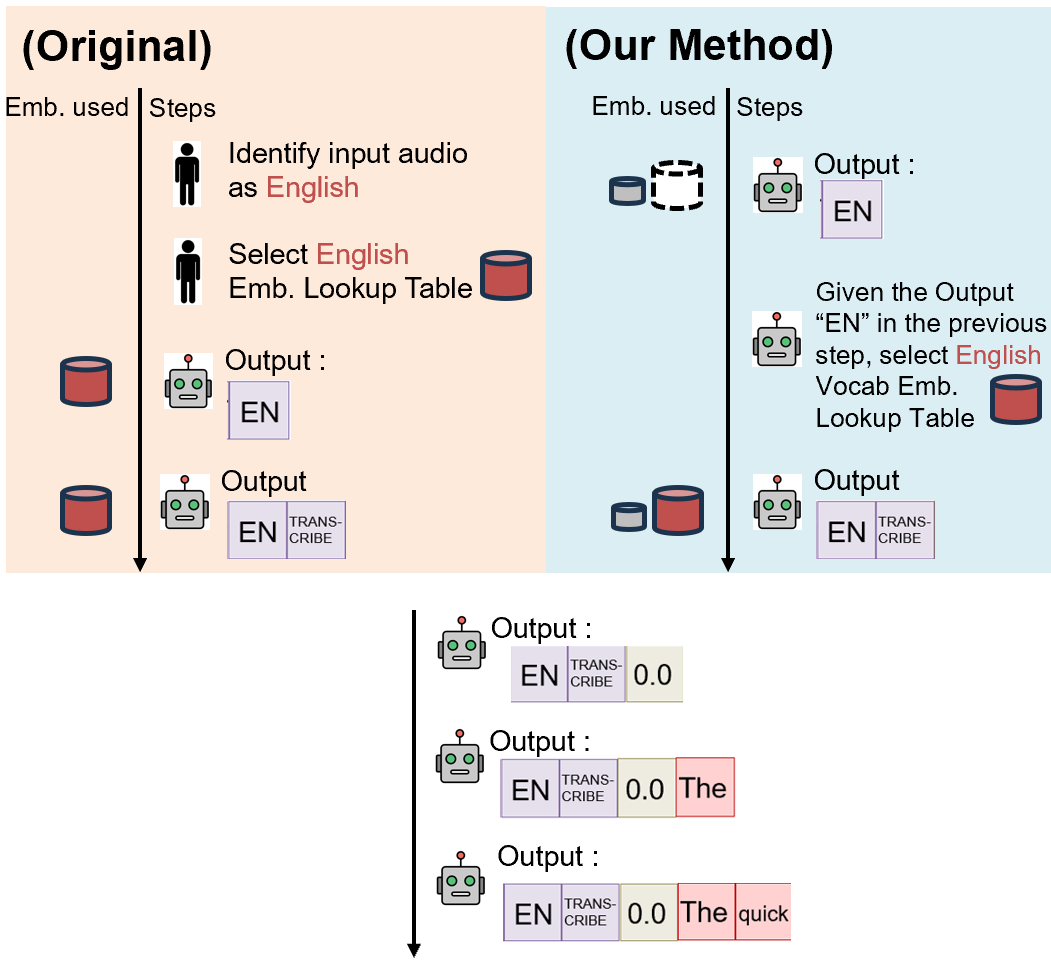}
  \caption{Overview of our language-agnostic decoding pipeline. Left: Manual selection of the language-specific token embedding lookup table is required to transcribe in the correct language. Right: A MMASR model first leverage the language-shared special token embedding lookup table as described in Figure \ref{fig:split_emb} to identify the language of the input audio, then automatically select the language-specific vocabulary token embedding lookup table for transcription.}
  \label{fig:task_agnostic}
\end{figure}

\subsection{Experience Replay}
A naive approach to mitigate CF is to retrain a model from scratch on a combination of all the previously learnt tasks' datasets and the new task's dataset. However, this is computationally expensive and it is not always possible to acquire all the previous tasks' dataset. As such, Experience Replay (ER) methods aim to remind a model of previously learnt tasks by mixing only a small amount of the old tasks' data \cite{brignac2023improving}, instead of all the data, with the new task's dataset during adaptation. The goal is to minimize: \setlength\abovedisplayskip{10pt}
\begin{equation}
\label{eqn:er}
\mathbb{E}_{(x,y)\sim{}D_B} [\ell(y, f_{\theta}(x))] + \beta *\mathbb{E}_{(x,y)\sim{}D_{\hat{A}}}  [\ell(y, f_{\theta}(x))]
\end{equation}
\noindent where $D_B$ is the new task B dataset, $\ell$ is the classification loss, $f_{\theta}$ is the model function parameterized by $\theta$, $D_{\hat{A}}$ is a subset of the old task A dataset and $\beta$ is a hyper-parameter balancing the trade-off between the terms. Equation \ref{eqn:er} can be easily extended to settings with more than 1 old task by adding extra terms for each old task correspondingly.

\subsection{Separate Token Embedding}

The adaptation of the token embedding lookup table \cite{huang2021lookup} at the decoder is prone to forgetting \cite{arora2019does}. This is because Whisper shares its tokens between languages. When the model is adapted to new languages, the token embeddings \cite{ethayarajh2019contextual} are also updated with word semantic information of the new language. This may overwrite the old languages semantics and cause forgetting.

Specifically, a token embedding lookup table is a matrix $\bm{A} \in \mathcal{R}^{E\times{}U}$ where $E$ is the embedding dimension and $U$ is the number of tokens in the model's vocabulary. Given that a Transformer decoder takes in the embeddings of the $K$ previously decoded tokens to predict the next token and $\bm{V}=[\vec{v_1},...,\vec{v_K}]$ contains $K$ columns of $U$-dimension one-hot vector \cite{karani2018introduction} of the decoded tokens. Then the token embeddings input $Z \in \mathcal{R}^{E\times{}K}$ feeding to the Transformer decoder is:

\begin{equation}
Z=\bm{A}\times{}\bm{V}
\end{equation}

To mitigate forgetting in the token embeddings, we propose to create a separate copy of the token embeddings for each new language as shown in Figure \ref{fig:split_emb}B and keep the original token embeddings for the old languages, such that the new languages embeddings can be adapted without modifying the original ones. The token embeddings for a new langauge are stored in a separate lookup table $\hat{\bm{A}} \in \mathcal{R}^{E\times{}J}$ where $J<U$ is the number of tokens used by the new language. The proportion of Whisper's tokens used by each new language is shown in Table \ref{tab:token_coverage}

\definecolor{ao(english)}{rgb}{0.0, 0.5, 0.0}
\begin{table}[]
\centering
\caption{Proportion of Whisper's tokens that appear in the ten hours train set of each new language.}
\begin{tabular}{cccccc}
\toprule
%\rowcolor[HTML]{C0C0C0} 
 language & \textbf{rw}   & \textbf{eo}   & \textbf{kab} & \textbf{lg}    & \textbf{mhr} \\
coverage (\%) & 6.4 & 7.9 & 4.5 & 4.6   & 2.8 \\
\cmidrule{1-6}
 language & \textbf{ckb}  & \textbf{ab}   & \textbf{kmr} & \textbf{fy-NL} & \textbf{ia}  \\
coverage (\%) & 0.4 & 1.4 & 3.9 & 5.6 & 7.9 
\\ \bottomrule
\end{tabular}
\label{tab:token_coverage}
\end{table}

\subsection{Language-agnostic Dynamic Architecture}

Although using language-specific token embeddings reduces forgetting, it is not language-agnostic. As shown in Figure \ref{fig:split_emb}A, the ASR decoder takes in a sequence of token embeddings to predict the next token. Figure \ref{fig:split_emb}B shows that a different token embedding lookup table is used for different languges. We further propose to split the lookup table into a language-shared Special Token (ST) part and a language-specific vocabulary part as shown in \ref{fig:split_emb}C. This is becuase the top-left part of Figure \ref{fig:task_agnostic} shows that normally if a language-specific text embedding layer is used, manual selection of the language-specific lookup table is required before transcription and the MMASR system is not language-agnostic. The right part of the figure shows that if the lookup table is split, the model can first perform LID using the language-shared ST part and then automatically select the language-specific vocabulary part for transcription. As such, the decoding is language-agnostic as the model can infer the language label by itself.

\begin{figure}[t]
  \centering
  \includegraphics[width=\linewidth]{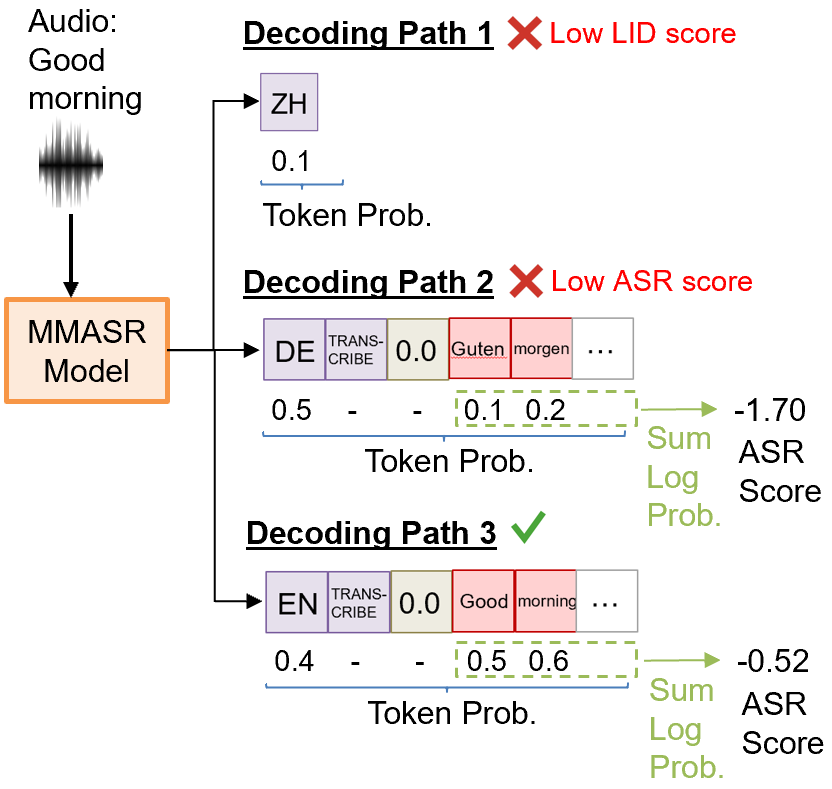}
  \caption{Overview of our task-wise beam search. During Decoding, the MMASR model first predicts the language ID of the input audio and decode the audio in the top $N=2$ scoring languages to generate $N$ hypothesis. Finally, the hypothesis with the highest ASR score is selected as the ASR output.
}
  \label{fig:task_beam_search}
\end{figure}

\subsection{Task-wise beam search}
The language-agnostic ASR performance highly relies on the accuracy of LID \cite{mishraend,liself,vachhani2023multi,praveen2023language} as wrong LID predictions may lead to transcription in the wrong language. To improve the LID and the language-agnostic ASR performance, we propose task-wise beam search, where we retain a list of candidates of different languages during beam search \cite{10.1007/978-3-031-70566-3_7} to improve its diversity \cite{vijayakumar2016diverse}.

Specifically as shown in Figure \ref{fig:task_beam_search}, an attempt to decode an audio in three languages is shown in decoding paths (DP) 1-3 respectively. First, the model performs LID by generating an LID token. Then DP1 is pruned as the LID token ``ZH" gives a low score of $0.1$. Only languages DE and EN, which have the top $N=2$ LID scores are kept and transcribed in DP2 and DP3. Next, the log probabilities of the transcribed tokens are summed to produce an ASR score for each language. Finally, the hypothesis in DP 3 is preferred over DP2 as its ASR score is higher.

In addition, we empirically find that Whisper may sometimes output blank hypothesis or output hypothesis in a different language than the one specified in the model's prompt. To increase the stability of our decoding method, it is important that we disable the task-wise beam search for a specific audio if we find that a decoding path has fewer than $M_{\rm len}$ words, or has more than $M_{\rm overlap}$ overlapped words with other paths.

\section{Experiment and Discussion}
\label{sec:exp-con}

\definecolor{ao(english)}{rgb}{0.0, 0.5, 0.0}
\begin{table*}[h!]
\centering
\caption{WER of adapting Whisper-small to new languages Esperanto (eo) and Interlingua (ia), and test forgeting on pretrained language English (en) and Germen (de) respectively. Language-agnostic \cite{datta2020language} results are obtained without manually specifying the language to transcribe, and vice versa for language-aware results.}
\begin{tabular}{ccccccccccc}
\toprule
%\rowcolor[HTML]{C0C0C0} 
\multirow{2}{1cm}{\textbf{Method}}                                            & \multicolumn{5}{c|}{\textbf{Language-aware WER (\%)}} & \multicolumn{5}{c}{\textbf{Language-agnostic WER (\%)}}\\
&  \textbf{en} & \textbf{de} & \textbf{eo} & \textbf{ia} & \multicolumn{1}{c|}{\textbf{avg}} & \textbf{en} & \textbf{de} & \textbf{eo} & \textbf{ia} & \textbf{avg}\\ \cmidrule{1-11} %\hline
Unadapted                                                                                                                                                              &             14.57 &  14.00 & n/a & n/a & \multicolumn{1}{c|}{n/a} & 14.57 & 14.00 & n/a & n/a & n/a        \\
FT                                                                                                                                                                &             68.99 & 64.83 & 18.04 & 12.31 & \multicolumn{1}{c|}{$41.0_{\color{gray}{-00.0\%}}$} & 90.55 & 88.40 & 18.04 & 12.31 & $52.3_{\color{gray}{-00.0\%}}$      \\ \cmidrule{1-11} %\hline
\textit{CL baselines}\\
AVG \cite{eeckt2022weight}                                                                                                                                                              &             16.20 & 17.94 & 38.96 & 16.53 & \multicolumn{1}{c|}{$22.4_{\color{ao(english)}{-45.4\%}}$} & 18.37 & 26.55 & 40.00 & 16.53 & $25.4_{\color{ao(english)}{-51.4\%}}$      \\
LwF \cite{li2017learning}                                                                                                                                                              &             16.04 & 16.81 & 22.20 & 14.82 & \multicolumn{1}{c|}{$17.5_{\color{ao(english)}{-57.3\%}}$} & 74.96 & 40.07 & 22.18 & 14.82 & $38.1_{\color{ao(english)}{-27.2\%}}$      \\
EWC \cite{kirkpatrick2017overcoming}                                                                                                                                                              &             15.62 & 23.78 & 19.21 & 13.31 & \multicolumn{1}{c|}{$18.0_{\color{ao(english)}{-56.1\%}}$ } & 39.87 & 41.12 & 19.21 & 13.34 & $28.4_{\color{ao(english)}{-45.7\%}}$       \\
MAS \cite{aljundi2018memory}                                                                                                                                                              &             15.64 & 18.29 & 20.14 & \textbf{12.42} & \multicolumn{1}{c|}{$16.7_{\color{ao(english)}{-59.3\%}}$} & 88.91 & 21.97 & 20.14 & \textbf{12.46} & $35.9_{\color{ao(english)}{-31.4\%}}$      \\
A-GEM \cite{chaudhry2018efficient}                                                                                                                                                            &             18.15 & 18.10 & \textbf{18.74} & \textbf{12.42} & \multicolumn{1}{c|}{$16.9_{\color{ao(english)}{-58.9\%}}$} & 19.09 & 18.68 & \textbf{18.91} & 12.68 & $17.3_{\color{ao(english)}{-66.9\%}}$      \\
DER \cite{buzzega2020dark}                                                                                                                                                               &             17.13 & 16.31 & 19.54 & 13.66 & \multicolumn{1}{c|}{$16.7_{\color{ao(english)}{-59.3\%}}$} & 56.95 & 23.59 & 20.25 & 13.66 & $28.6_{\color{ao(english)}{-45.3\%}}$      \\ 
ER \cite{brignac2023improving}                                                                                                                                                               &             15.64 & 16.65 & 20.34 & 12.94 & \multicolumn{1}{c|}{$16.4_{\color{ao(english)}{-60.0\%}}$} & 18.26 & 17.20 & 20.45 & 13.14 & $17.3_{\color{ao(english)}{-66.9\%}}$      \\ \cmidrule{1-11} %\hline
\textit{our methods}\\
\begin{tabular}[c]{@{}l@{}}ER-E\end{tabular}                                          &             \textbf{14.48} & \textbf{14.34} & 20.32 & 13.58 & \multicolumn{1}{c|}{$\mathbf{15.7}_{\color{ao(english)}{-61.7\%}}$} & 17.97 & 14.98 & 20.49 & 13.97 & $16.9_{\color{ao(english)}{-67.7\%}}$      \\
\begin{tabular}[c]{@{}l@{}}ER-E-B\end{tabular}                                          &             \textbf{14.48} & \textbf{14.34} & 20.32 & 13.58 & \multicolumn{1}{c|}{$\mathbf{15.7}_{\color{ao(english)}{-61.7\%}}$} & \textbf{14.83} & \textbf{14.69} & 20.34 & 13.55 & $\mathbf{15.9}_{\color{ao(english)}{-69.6\%}}$      \\
\bottomrule
\end{tabular}
\label{tab:main_result}
\end{table*}

\subsection{Dataset and Model Details}
We implement our CL methods based on the popular SpeechBrain \cite{ravanelli2021speechbrain} toolkit and CL-MASR \cite{della2023cl}.

%\footnote{\url{https://commonvoice.mozilla.org/en}}
Following previous works \cite{yu2023master,della2023cl},  we evaluate our method on a subset of the widely used large-scale CommonVoice dataset \cite{ardila2019common}. We follow CL-MASR \cite{della2023cl} to extract the data subsets. They consists of ten languages unseen by Whisper and ten seen languages. Each language contains 10 hours of data for training, 1 for validation, and 1 for testing. For Chinese, traditional Chinese characters are converted to simplified Chinese. We adapt small and large-v2 variants of Whisper in two CL settings: 1) Adapt to one unseen language and test forgetting on one seen language, 2) Adapt to ten unseen languages sequentially and test forgetting on ten seen languages, and a maximum of six days are required utilizing an NVIDIA A40 GPU.

For each unseen language, we adapt the 2 variants of Whisper for 2 epochs with a train batch size of 4. Only the weights of the Whisper decoder is updated and the encoder is frozen \cite{della2023cl}. For ER, the replay data size is one hour for each new language. We use AdamW \cite{loshchilov2017decoupled} as the optimizer and a variant\footnote{\url{https://speechbrain.readthedocs.io/en/latest/\_modules/speechbrain/nnet/schedulers.html\#NewBobScheduler}} of the ReduceLROnPlateau\footnote{\url{https://pytorch.org/docs/stable/generated/torch.optim.lr\_scheduler.ReduceLROnPlateau.html}} learning rate (LR) scheduler. Validation is done at an interval of $1/32$ epoch. We sweep through the hyper-parameters to tune them for all methods.

We refer to our methods as 1) ER-E-B, which is ER with separate token embeddings and task-wise beam search, and 2) ER-E, which is ER with separate token embeddings only. We set $N=2$, $M_{len}=5$ and $M_{overlap}=3$.

\subsection{Results and Discussion}
\begin{figure}[h!]
  \centering
  \includegraphics[width=\linewidth]{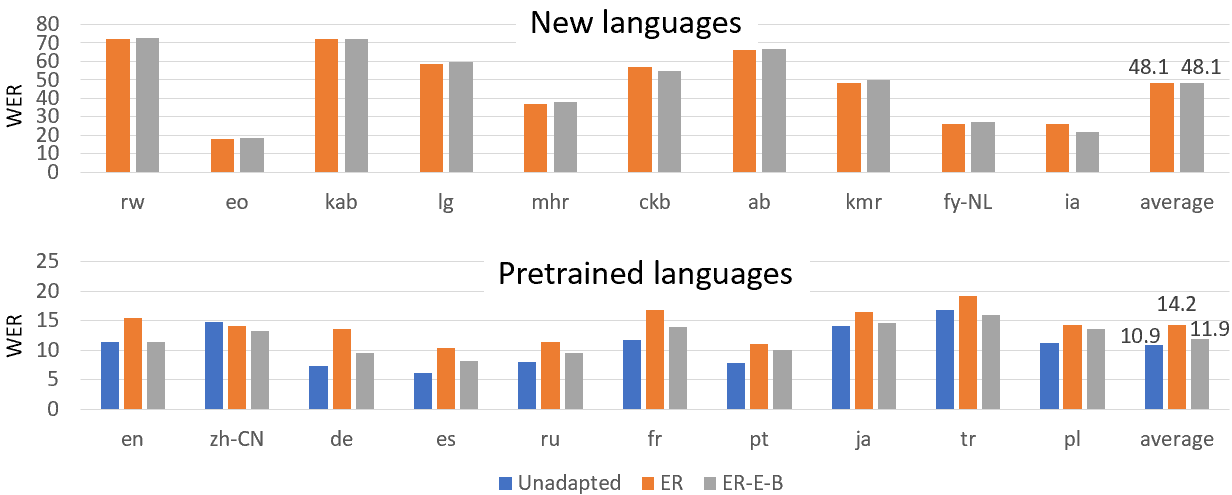}
  \caption{WER of adapting whisper-large-v2 sequentially to 10 new languages of varying difficulties and test forgetting on 10 pretrained languages in the language-agnostic setting. ``Unadapted" means the unadapted model.}
  \label{fig:wer}
\end{figure}

We present the results for both whisper-small and whisper-large-v2 models. Table \ref{tab:main_result} shows the result of adapting Esperanto and Interlingua for whisper-small. Full fine-tuning (FT) can improve WER for the target languages, but incurs CF. In contrast, all Continual Learning (CL) baseline methods significantly reduce CF, reducing AWER by 27.2-66.9\% relatively compared with FT. Among these baseline methods, ER performs better in overall performance, though EWC and MAS has better results for English and Esperanto for language-aware results. However, AVG performs less well relative to ER, as task-specific layers are not used as in \cite{eeckt2022weight}.

Results show that our methods outperform all CL baselines and improve upon FT result by $61.7-69.6\%$. Forgetting is further mitigated compared with all CL baselines while our method can maintain similar WER for the newly adapted languages. Furthermore, our ER-E-B method has almost no forgetting in English and German even in the more difficult language-agnostic setting.

Lastly, we adapt whisper-large-v2 sequentially to the 10 new languages of varying difficulties and plot the results in Figure \ref{fig:wer}. Our method ER-E-B outperforms the best CL baseline ER and reduces the AWER of pretrained languages from $14.2\%$ to $11.9\%$ without compromising the AWER of new languages. We do not compare with dynamic-architecture-based methods as they are not language-agnostic.

% Extremely long decoding

\subsection{Ablation study}

We further perform ablation studies for Task-wise Beam Search and Separate Token Embedding. For the experiments, we adapt whisper-small to a new language Esperanto (eo) and test forgetting on a pretrained language English (en). Table \ref{tab:ablation} shows that for Task-wise Beam Search, it consistently improves performance for ER and ER-E methods in the language-agnostic setting by up to a relative $8.3\%$ AWER. 

\begin{figure}[h!]
  \centering
  \includegraphics[width=0.9\linewidth]{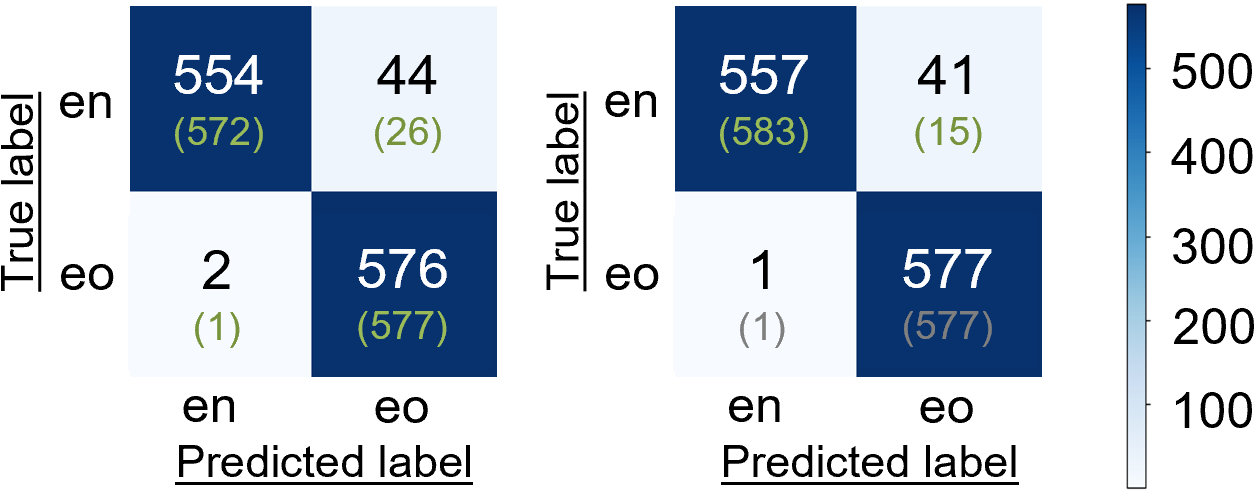}
  \caption{LID Confusion matrix on Esperanto (eo) and English (en) using Whisper-small adapted with ER (left) and ER-E (right). The results after adding task-wise beam search are shown in parenthesis.}
  \label{fig:confusion_matrix}
\end{figure}

We hypothesize that the above improvement is caused by the decrease in LID errors. To verify this, we therefore plot the LID confusion matrix for the above experiments. As shown in Figure \ref{fig:confusion_matrix}, most LID errors in the language-agnostic setting originate from incorrectly identifying English audios as Esperanto. Task-wise beam search reduces such errors by more than $40\%$ for ER and $60\%$ for ER-E.

\definecolor{ao(english)}{rgb}{0.0, 0.5, 0.0}
\begin{table}[]
\centering
\caption{Ablation study of Task-wise Beam Search.}
\begin{tabular}{cccc}
\toprule
%\rowcolor[HTML]{C0C0C0} 
\multirow{2}{3.5cm}{\textbf{Method}} & \multicolumn{3}{c}{\textbf{WER (\%)}}\\
& \multirow{1}{*}{\textbf{en}} & \multirow{1}{*}{\textbf{eo}} & \multirow{1}{*}{\textbf{avg}} \\ \cmidrule{1-4}%\hline
\multirow{1}{3.5cm}{ER} & \multirow{1}{*}{18.26} & \multirow{1}{*}{20.45} & \multirow{1}{*}{19.4}  \\
\multirow{1}{3.5cm}{\footnotesize{\quad + Task-wise Beam Search}} & \multirow{1}{*}{16.34
} & \multirow{1}{*}{20.32} & \multirow{1}{*}{18.3}
\\ \cmidrule{1-4}
\multirow{1}{3.5cm}{ER-E} & \multirow{1}{*}{17.97} & \multirow{1}{*}{20.49} & \multirow{1}{*}{19.2}  \\ 
\multirow{1}{3.5cm}{\footnotesize{\quad + Task-wise Beam Search}} & \multirow{1}{*}{14.83} & \multirow{1}{*}{20.34} & \multirow{1}{*}{17.6}
\\ \bottomrule
\end{tabular}
\label{tab:ablation}
\end{table}

Additionally, Table \ref{tab:ablation2} shows that adding Separate Token Embedding to ER reduces AWER from $16.4$ to $15.7$. Most of the reduction comes from reducing WER in English and German. We hypothesize that sharing sub-word token embeddings between languages causes forgetting in the languages, and using a language-specific embedding layer can mitigate it. As an alternative, we question whether only updating the sub-word token embeddings that appeared in the new language dataset and freezing the other embeddings can also reduce forgetting. We call this method ER-E-part. The results show that ER-E part can also reduce forgetting to some degree. However, we emphasize that ER-E-part only works when the token set of the new language dataset does not cover too much of the pre-trained model token set, while ER-E does not have such limitation.

\definecolor{ao(english)}{rgb}{0.0, 0.5, 0.0}
\begin{table}[]
\centering
\caption{Ablation study of Separate Token Embedding.}
\begin{tabular}{cccccc}
\toprule
%\rowcolor[HTML]{C0C0C0} 
\multirow{2}{1.4cm}{\textbf{Method}} & \multicolumn{5}{c}{\textbf{WER (\%)}}\\
& \multirow{1}{*}{\textbf{en}} & \multirow{1}{*}{\textbf{de}} & \multirow{1}{*}{\textbf{eo}} & \multirow{1}{*}{\textbf{ia}} & \multirow{1}{*}{\textbf{avg}} \\ \cmidrule{1-6}%\hline
\multirow{1}{1.4cm}{ER-E} & \multirow{1}{*}{14.48} & \multirow{1}{*}{14.34} & \multirow{1}{*}{20.32} & \multirow{1}{*}{13.58} & \multirow{1}{*}{15.7}  \\
\multirow{1}{1.5cm}{ER-E-part} & \multirow{1}{*}{14.96} & \multirow{1}{*}{14.56} & \multirow{1}{*}{19.88} & \multirow{1}{*}{14.12} & \multirow{1}{*}{15.9}  \\ 
\multirow{1}{1.4cm}{ER} & \multirow{1}{*}{15.64} & \multirow{1}{*}{16.65} & \multirow{1}{*}{20.34} & \multirow{1}{*}{12.94} & \multirow{1}{*}{16.4} 
\\ \bottomrule
\end{tabular}
\label{tab:ablation2}
\end{table}

\section{Conclusion}

% Conclusion
To conclude, we present ER-E-B, a CL method that outperforms ER methods in mitigating CF and ablation study has shown the effectiveness of our proposed language-agnostic dynamic architecture, task-wise beam search and separate token embedding.

\vfill\pagebreak

\bibliographystyle{IEEEtran}%\footnotesize
\bibliography{paper}

\end{document}